\documentclass[a4paper, 10pt, conference]{IEEEtran}
\IEEEoverridecommandlockouts
\usepackage{graphicx, lipsum}
\usepackage{subcaption}
\usepackage{multicol}
\usepackage{multirow}
\usepackage{array}
\usepackage{pgfplots}
\pgfplotsset{compat=1.7}
\usepackage{graphicx}
\usepackage{amsmath}
\usepackage{amssymb}
\usepackage{booktabs}
\usepackage{mathtools}
\usepackage{float}
\usepackage{placeins}

\usepackage{tikz}
\usepackage{caption}
\usepackage{setspace}
\usepackage{parskip}
\usepackage{xcolor}

\usepackage{cite}
\usepackage{amsmath,amssymb,amsfonts}
\usepackage{algorithmic}
\usepackage{graphicx}
\usepackage{textcomp}
\usepackage{xcolor}
\def\BibTeX{{\rm B\kern-.05em{\sc i\kern-.025em b}\kern-.08em
    T\kern-.1667em\lower.7ex\hbox{E}\kern-.125emX}}
\begin{document}

\title{Do we need entire training data for adversarial training?
}

\author{\IEEEauthorblockN{Vipul Gupta}
\IEEEauthorblockA{\textit{Department of Computer Science} \\
\textit{Pennsylvania State University}\\
vkg5164@psu.edu}
\and
\IEEEauthorblockN{Apurva Narayan}
\IEEEauthorblockA{\textit{Department of Computer Science} \\
\textit{University of British Columbia}\\
apurva.narayan@ubc.ca}
}

\maketitle

\begin{abstract}
Deep Neural Networks (DNNs) are being used to solve a wide range of problems in many domains including safety-critical domains like self-driving cars and medical imagery. DNNs suffer from vulnerability against adversarial attacks. In the past few years, numerous approaches have been proposed to tackle this problem by training networks using adversarial training. Almost all the approaches generate adversarial examples for the entire training dataset, thus increasing the training time drastically. We show that we can decrease the training time for any adversarial training algorithm by using only a subset of training data for adversarial training. To select the subset, we filter the adversarially-prone samples from the training data. We perform a simple adversarial attack on all training examples to filter this subset. In this attack, we add a small perturbation to each pixel and a few grid lines to the input image.

We perform adversarial training on the adversarially-prone subset and mix it with vanilla training performed on the entire dataset. Our results show that when our method-agnostic approach is plugged into FGSM \cite{b9}, we achieve a speedup of 3.52x on MNIST and 1.98x on the CIFAR-10 dataset with comparable robust accuracy. We also test our approach on state-of-the-art Free adversarial training \cite{b24} and achieve a speedup of 1.2x in training time with a marginal drop in robust accuracy on the ImageNet dataset.
\end{abstract}

\begin{IEEEkeywords}
Adversarial Training, Adversarial Robustness, Adversarial attacks
\end{IEEEkeywords}

\section{Introduction}

% \vipul{Para1} DNNs are applied in various applications. Safety critical tasks require them to trustworthy.
In the past few years, Deep Neural Networks (DNNs) have shown the state of the art performance on a wide range of domains like computer vision \cite{b2}, reinforcement learning \cite{b56} and natural language processing \cite{b3}. DNNs are now being applied in various domains such as credit scoring \cite{b4}, judicial system \cite{b1}, self driving cars \cite{b5}, and medical imagery \cite{b6}. Application in these domains requires DNNs to be trustworthy and dependable \cite{b7, b8}.

% \vipul{Para2} DNNs are prone to adversarial attacks. Recently a lot of approaches have been proposed for adversarial training.
On the other hand, DNNs offer poor robustness against adversarial attacks \cite{b9, b10}. A small, human imperceptible, perturbation in test input can easily fool the neural network and result in a significant change in the output of the network \cite{b12, b11, b55}. In a safety-critical application, the vulnerability against minor perturbations can be exploited by attackers to change the output significantly leading to catastrophic outcomes \cite{b8}. Recently, numerous approaches have been proposed to improve the robustness of DNNs using adversarial training, a technique in which a network is trained on adversarial examples \cite{b13, b14, b15, b16}. Adversarial examples are small modifications to training input samples, which are carefully crafted to change model's prediction. 

% \vipul{Para3} Adversarial training is of 2 types - targeted and untargeted.
Adversarial training can be classified into 2 types: untargeted and targeted. In untargeted adversarial training, the aim is to find a small perturbation for the input to misclassify it to any incorrect class like FGSM \cite{b9}, Fast Gradient Value \cite{b17}, universal adversarial perturbations \cite{b18}, black-box attacks \cite{b19}. More recent techniques focus on target-specific adversarial training \cite{b20, b21, b14, b22, b23}. Here the aim is to find a small perturbation that will change the output of the input image to a particular target class.

% \vipul{Para4} One of the main issues with adversarial training is the training time. They add a lot of overhead.
One of the main issues with most of the proposed methods for adversarial training is the increase in training time as pointed in \cite{b25, b26}. Adversarial training adds a lot of overhead to vanilla training and increases the training time by many folds. The initial approaches like FGSM \cite{b9} take around 8-9x more time (on datasets like CIFAR-10) when compared with vanilla training. This gigantic increase in training time makes adversarial training not feasible for many real-world problems and applications. The high computation cost of adversarial training has motivated a lot of work \cite{b24, b25, b26, b57, b58, b59}, they aim to reduce the training time without compromising on the robustness accuracy.

% \vipul{Para5}
All the proposed methods for decreasing the training time for adversarial methods performs optimisation on models \cite{b24, b25, b26}. In this paper, we are first to show that we do not need to perform adversarial attacks on the entire training data for adversarial training. We show that we can achieve similar performance by using only a fraction of training data for adversarial example generation, thus decreasing the training time drastically. We propose a method-agnostic approach where we mix adversarial training with vanilla training. 

% \vipul{Para6} 

We perform a simple adversarial attack on all training samples, to filter the adversarially-prone subset from the training set. In this attack, we add a small perturbation to each pixel and a few grid lines to the input image. The adversarially-prone subset contains samples that are more prone to adversarial attacks than other samples in the training set. We perform adversarial training only on the adversarially-prone subset.
% will use only adversarially-prone data for adversarial training.

% \vipul{Para7} 
We perform extensive experiments on the famous, FGSM approach \cite{b9}. We show the performance on CIFAR-10 and MNIST datasets. By combining vanilla training with adversarial training on the adversarially-prone subset, we reach comparable robustness with 2 to 3.5 times faster training time. We also apply our approach to the state-of-the-art, Free adversarial training \cite{b24} method on the ImageNet dataset. We show a speedup of 1.2 times with decreasing the robust accuracy marginally by 1.68\% on Free adversarial training. 

Our main contributions in this paper are two-fold. First, we show that we do not need entire training dataset for adversarial training. Second, we propose a method to filter adversarially-prone subset of training data and perform adversarial training only on this subset. Our proposed approach speeds up training time for any adversarial training method and achieving comparable accuracy.

% We thus argue that we don't need the entire training dataset to generate adversarial samples for robustness. We present a method-agnostic plug-and-play tool, that can be used with any existing adversarial training algorithm.
% Talk about results

\begin{figure}
     \begin{subfigure}[b]{0.145\textwidth}
         \includegraphics[width=2.5cm, height=2cm]{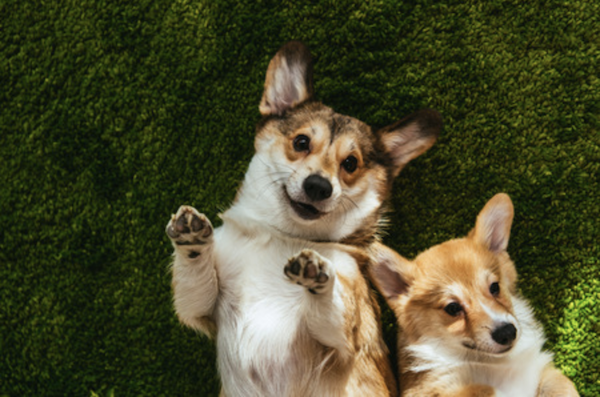}
         \caption{Pembroke - 81.2\%\\Chihuahua - 9.3\%}
         \label{fig: raw }
     \end{subfigure}
     \hspace{0.2cm}
     \begin{subfigure}[b]{0.145\textwidth}
         \includegraphics[width=2.5cm, height=2cm]{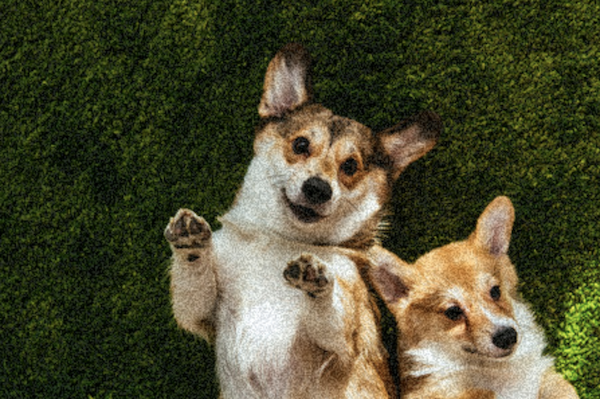}
         \caption{Pembroke - 15.4\%\\Chihuahua - 57.9\%}
         \label{fig: p1}
     \end{subfigure}
     \hspace{0.2cm}
     \begin{subfigure}[b]{0.145\textwidth}
         \includegraphics[width=2.5cm, height=2cm]{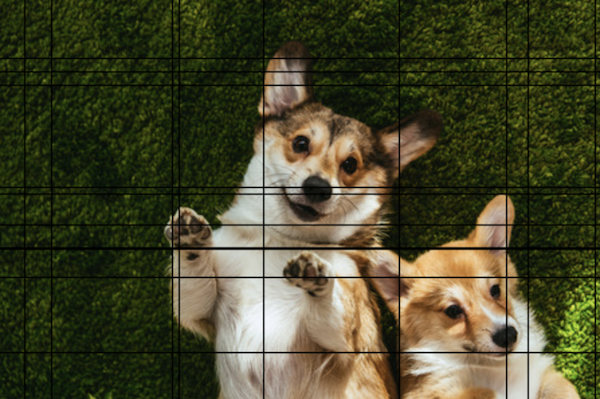}
         \caption{Pembroke - 33.6\%\\Chihuahua - 54.6\%}
         \label{fig: p2}
     \end{subfigure}
        \caption{A sequence showing effect of small perturbation to input image. (a)original image (b)every pixel value in input is modified by a random number between +30 to -30 (c)few random lines are added to the original image}
        \label{fig: p3}
\end{figure}

\section{Related Work}

\paragraph{Adversarial Attacks and Training} Adversarial examples for deep neural networks were first discovered by \cite{b11}. One of the most popular adversarial attacks, FGSM \cite{b9} perturbs every input dimension but with a small quantity in the direction of the sign of the gradient. A lot of approaches have been proposed to generate adversarial attacks \cite{b20, b27, b28, b13, b31}. Some approaches work on changing a single pixel \cite{b33}, while some make minimal modification to input image such as blurring \cite{b34}, adding a small sub-network \cite{b37}, spatial transformation \cite{b38}. Some attacks have show strong transferability of adversarial examples across neural network models \cite{b39}. In \cite{b34} authors show that approaches like simple blurring and modifying a single optimal pixel by a large amount are able to fool the network. 

In \cite{b32} authors show that by modifying all pixels in the input image by a random number between 30 to 60, they were able to reduce confidence for 65.9\% images and generate 44.7\% misclassifications on different datasets. \cite{b17} provides a way to generate multiple perturbations for a single image. \cite{b18} provides an image-agnostic approach for generating perturbations. Some works have been proposed on evaluating the robustness against strong attacks \cite{b40, b41, b42, b43}. Advances in adversarial attacks are complemented by advances in adversarial training \cite{b16, b14}. \cite{b44} uses high level features to guide the pixel denoiser, \cite{b45} proposed an approach that works with large datasets networks, \cite{b46} uses randomized smoothing, \cite{b47} uses training data distribution to perturb images. 

Recent work on adversarial training aims at reducing the overhead added by training by adversarial examples \cite{b48, b49, b50}. \cite{b24} recycles gradient information during the model update for faster example generation, \cite{b25} uses random initialization to further speed up adversarial training. \cite{b26} trains a feed-forward neural network in a self-supervised manner to generate adversarial examples.

\paragraph{Representing inputs as intervals} Interval analysis has been used in various different approaches. \cite{b51, b52} provides a certifiable robustness for small perturbations, \cite{b53} find adversarial perturbation by formulating it as mixed integer programming, \cite{b29, b30} used interval analysis to find tight bounds on DNN outputs for safety purposes. \cite{b54} used it to find minimum distortion in input to generate adversarial examples. In this work, we will be using input interval to filter adversarially prone examples from the training dataset.

\section{Method}

\begin{figure}
    \centering
    \includegraphics[width=1.0\linewidth]{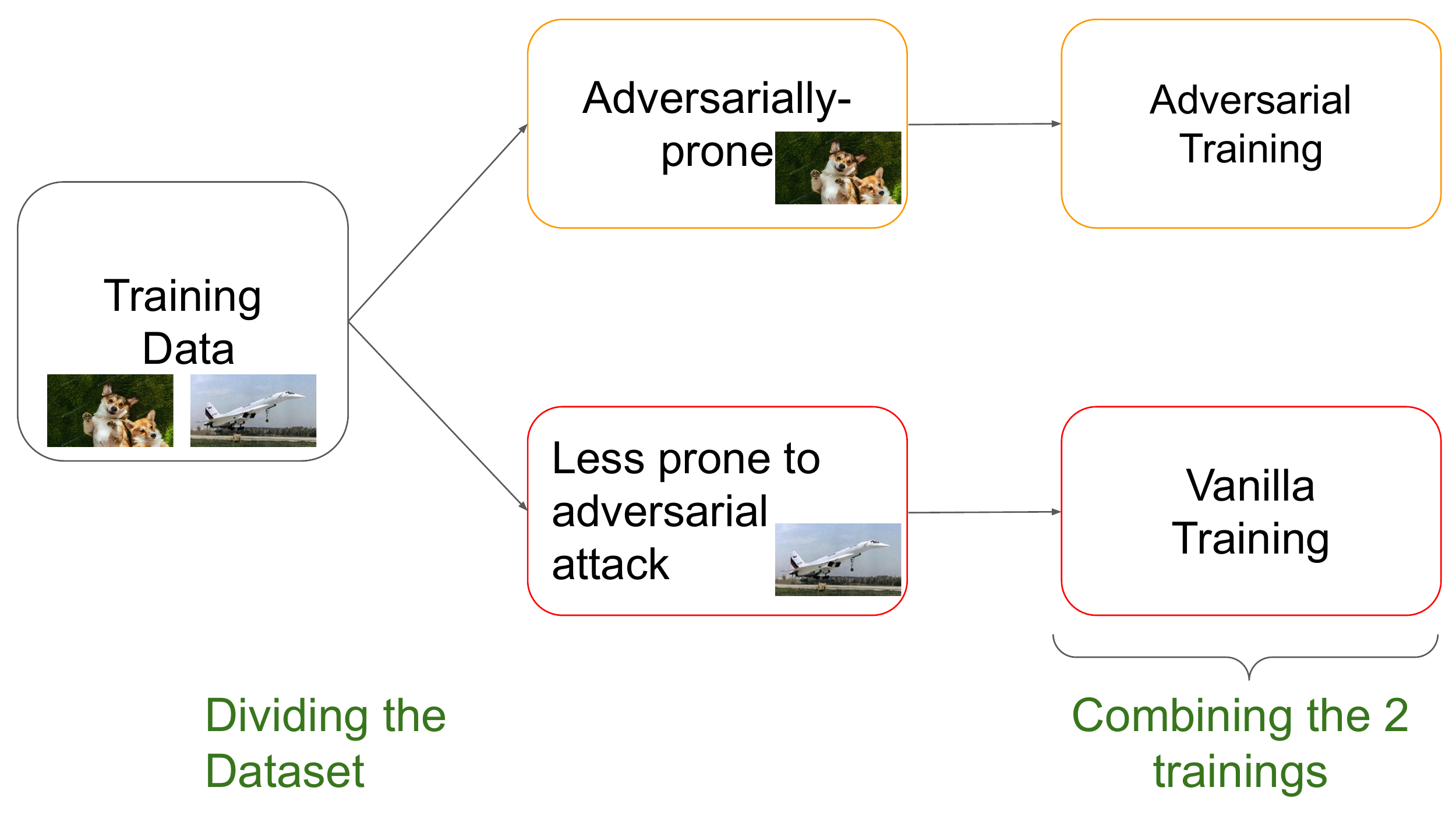}
    \caption{Overview of our method. We first divide the training dataset into 2 parts based on susceptibility to adversarial attacks. Then we perform adversarial training only on adversarially-prone subset and combine it with vanilla training. 
    % Training time can be decreased by this technique with a marginal decrease in model accuracy.
    }
    \label{fig:method_fig}
\end{figure}

\begin{figure*}
     \centering
     \begin{subfigure}[b]{0.25\textwidth}
         \centering
         \includegraphics[width=2.4cm, height=4.2cm]{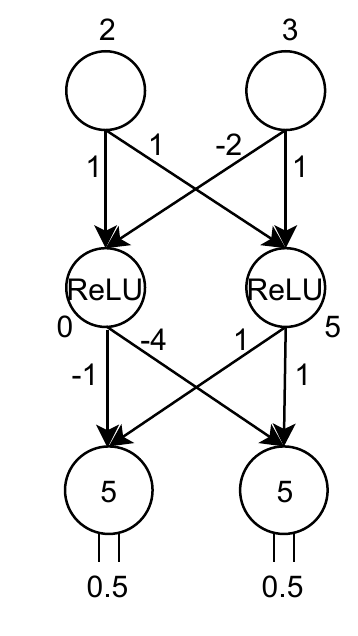}
         \caption{Raw input}
         \label{fig: raw}
     \end{subfigure}
     \hfill
     \begin{subfigure}[b]{0.25\textwidth}
         \centering
         \includegraphics[width=2.8cm, height=4.2cm]{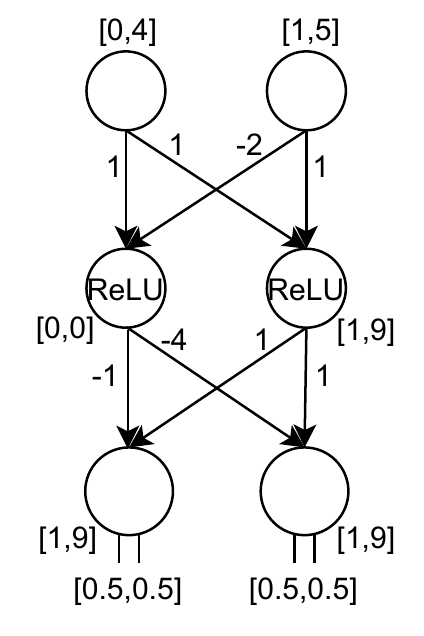}
         \caption{Input as interval}
         \label{fig: interval}
     \end{subfigure}
     \hfill
     \begin{subfigure}[b]{0.25\textwidth}
         \centering
         \includegraphics[width=2.4cm, height=4.2cm]{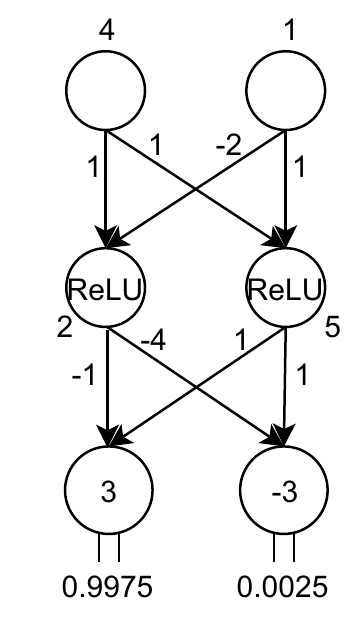}
         \caption{Random values within interval}
         \label{fig: combination}
     \end{subfigure}
        \caption{Illustration showing (a)propagation of raw input to the network, (b) changing the input to have an interval of length 2 and propagating upper and lower bounds (0,1), (4,5) (c)propagating random values from the input interval. This shows that we can get extreme values of an output range for any random input within the input interval} 
        \label{fig: three graphs}
\end{figure*}

% \vipul{Para1} Define vanilla training and adv training. Overview of the method. Filtering data, using only adv-prone data. Interval analysis.
In this work, we refer to vanilla training as normal training without adversarial examples. Adversarially-prone samples are samples which are more prone to adversarial attacks as compared to other samples. We aim to find an adversarially-prone subset of training data.
% We aim to find an adversarially-prone subset of training data, i.e. training samples which are more prone to adversarial attacks as compared to other samples. 
To achieve this, we use the concept of interval analysis as proposed by \cite{b29} to find a prediction range for each training sample.

% \vipul{Para2} We do not modify the way to generate adversarial attacks during adversarial training. Combine vanilla training and adversarial training.
% After filtering the adversarially-prone subset, we perform adversarial training only on this subset. 
Our method-agnostic approach can be used with any adversarial training method. We do not modify the way adversarial examples are being generated by the adversarial training method. Our approach proposes to perform adversarial training only on adversarially-prone subset rather than the entire training dataset.
% We do not modify the way to generate adversarial examples and the generation depends on the attack method being used. This makes it a method-agnostic approach. 
We combine vanilla training with adversarial training in a ratio of 2:1, thus performing adversarial training on every third iteration.

\subsection{Filtering adversarially-prone data}
% \vipul{Para3} Define interval analysis and give details.
\cite{b32} showed that perturbation of 60 pixels gives the best results for adversarial attacks. We used results from \cite{b32} to select the interval length of perturbation to be 60 pixels. This perturbation interval is used for filtering the adversarially-prone subset and each pixel in the input image is perturbed by a maximum of 60 pixels. For example, let's say that the original pixel value is x, the perturbation interval will be [x-60, x+60]. We aim to find the model prediction range for the image in this perturbation interval. 

Our aim is to find the model prediction range across the perturbation interval with minimum computation required. Initial assumption would be that we can get prediction range by measuring model's prediction at extreme ends of the prediction range. We show that this assumption is wrong and that extreme values of the prediction range can exist for any set of inputs.

Let us take the example of Figure \ref{fig: three graphs}. We have a single hidden layered neural network with ReLU activation function in the first layer and softmax at the last layer and weights of nodes are selected as shown in the figure. Propagation of inputs 2 and 3 through the network, produces an output of 0.5 for each class. Now let's consider a perturbation of length 2 pixels. So now the inputs can be taken anywhere between [0,4] and [1,5] respectively. In figure 1(b), the lower and upper bounds of the perturbation interval are propagated i.e {0,1} and {4,5}. The network outputs 0.5 for both these cases. In Figure 1(c), we select random value {4,1} for propagation. The network outputs 0.9975 for class 0 and 0.0025 for class 1. This shows that we can get extreme values for the prediction range for any set of inputs.
% This shows that the output of the network varies to an extent of misclassification, with minor variations in input. Since we don't have any prior knowledge about perturbation, any set of input values can give us extreme values for the output range. One main objective is to maximize the output variability with a minimum amount of computation effort.
% \vipul{Para4} We have used 2 types of adversarial attacks to filter adversarially-prone data. We used these 2 because..

\cite{b34} shows that blurring the image changes model's prediction for a lot of samples. \cite{b35} shows the importance of randomisation in adversarial attacks. \cite{b36} shows adding random lines in the image changes models prediction for majority of samples. All these approaches takes very less computation time and thus we use the works from \cite{b36, b35, b32, b34}, to perform 2 types of adversarial attacks to determine the prediction range. 

In the first attack, we perturb every pixel value by a random number between 0 and 60. In the second attack, we add single-pixel grid lines along the height and width at randomly selected rows and columns. We perform both these attacks on the input to get a prediction range. We choose these two attacks as they take very little computation time and result in a good adversarially-prone subset. We tried adding more attacks but they increased computation time with a small increase in the size of the adversarially-prone subset, so we decided to use these two attacks for our approach.

% \vipul{Para5} Details of the 2 attacks. Perturbation value, number of grid lines etc
We perform these attacks 3 times to get a prediction range for each sample. As the model trains, the adversarial boundaries keep changing. To get the updated adversarially-prone subset, we filter this subset after every 4 epochs. Using the prediction range, we filter samples based on the change in the highest probability prediction class across the prediction range. For the first type of attack, we select a random pixel value between -60 to +60 for each pixel. For the grid lines attacks, we select min(5, 0.05*height) and min(5, 0.05*width) as the number of grid lines for height and width respectively.

\subsection{Combining adversarial and vanilla training}

% \vipul{Para6} Details on how we combine the 2 training ways. Every 3 iterations.
We combine adversarial training and vanilla training by performing these 2 training procedures alternatively. From our analysis, we found that using a ratio of 2:1 between vanilla and adversarial training gave us the optimal results. Thus we perform adversarial training on every 3rd iteration and vanilla training on the remaining iterations. This ensures that the models see a mix of normal training samples and adversarial examples generated by using the most adversarially prone sample from the dataset. This helps in decreasing the training time as compared to general adversarial training, with a marginal decrease in accuracy. More analysis on this can be found in the experiments section. 

% \vipul{Para7} We update the adv-prone data every 4 epochs. This is ensure that we are getting the most adv-prone data as the model is training.

% \vipul{Para8} This method determines samples that are close to decision boundary and uses only that for adv-training.
Our approach to select the adversarially-prone subset helps us to find the training samples that are close to the decision boundary of a class learned by the model. Performing a small perturbation in the input changes the predicted class for these samples. Thus for adversarial training, we want to use only the training samples which are close to the decision boundary and thus are more prone to adversarial attacks. This helps in decreasing the training time without affecting the model performance. Our proposed approach uses only about 30-40\% of the training dataset for adversarial training.

Previous work which focus on improving training time \cite{b24, b26} does improvement on gradient update or way of training. To best of our knowledge, we are the first to propose using a subset of training set for adversarial training. When clubbed with other works on improving training time, our approach can further decrease the training time for any adversarial training method.

\section{Experiments}

\begin{table*}[h]
\centering
\begin{tabular}{l|l|c|c|c|c}
\toprule 
            % \multicolumn{4}{c}{MCAN}     \\ \hline
Method and dataset used                                                 &  Training Procedure     &  Vanilla Acc & Robust Accuracy  &  Train Time &  Improvement \\ \hline
\multirow{2}{*}{FGSM \cite{b9} on MNIST}        &   Adversarial           &    \multirow{2}{*}{99.21} & 97.71          &     37.9 min   &   \multirow{2}{*}{3.52}       \\ 
                                                                        &   Vanilla + Adv &                           & 98.71          &     10.8 min   &          \\ \hline 
\multirow{2}{*}{FGSM \cite{b9} on CIFAR-10}     &   Adversarial           &    \multirow{2}{*}{97.50} & 86.10          &     49.7 hr    &   \multirow{2}{*}{1.98}       \\ 
                                                                        &   Vanilla + Adv &                           & 85.46          &     25.1 hr    &          \\ \hline 
\multirow{2}{*}{Free adv \cite{b24} on ImageNet}       
                                                                        &   Adversarial           &    \multirow{2}{*}{90.88} & 83.52          &     64.1 hr    &   \multirow{2}{*}{1.20}       \\ 
                                                                        &   Vanilla + Adv &                           & 81.84          &     53.5 hr    &          \\ \bottomrule 

\end{tabular}
\caption{We show the results of using our proposed technique on adversarial training techniques such as FGSM \cite{b9} and Free adversarial training \cite{b24}. First we train networks using the adversarial training and then we mix vanilla and adversarial training. Robust accuracy is the accuracy of the trained model towards adversarial attacks. Last column shows speedup in training time by using our proposed method. We show speed-up in training time with marginal change in robust accuracy.}
\vspace{-1.5em}
\label{tab:results_tab}
\end{table*}

In this section, we present experimental results of our novel strategy on 2 adversarial training algorithms : FGSM \cite{b9} and Free Adversarial training \cite{b24}. We show that we achieve comparable adversarial robustness, without changing the way of generating adversarial attacks with a significant decrease in training time. To the best of our knowledge, we are the first to present an optimization approach that can be used with any adversarial training algorithm. So direct comparison with other methods is not possible at this point. Our proposed method can have slightly lower robust accuracy than using only adversarial training as shown on CIFAR-10 and ImageNet in Table \ref{tab:results_tab}. Our method is useful in cases where significant decrease in training time is more useful a slight decrease in robust accuracy. 

For our analysis, we evaluate accuracy of models on adversarial attacks. We use robust accuracy as the measurement metric. Robust accuracy is the accuracy of the trained model towards adversarial attacks. We also report vanilla accuracy on each dataset, which is the accuracy without adversarial training. This column in table \ref{tab:results_tab} is just for comparing vanilla training with adversarial training. As expected the adversarial accuracy is lower than vanilla accuracy on every dataset.

% \vipul{Para1}

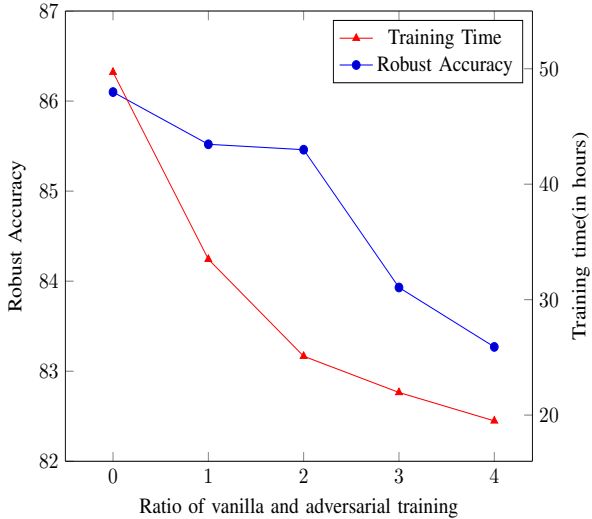
\begin{figure}
\resizebox{8cm}{7cm}
{
\begin{tikzpicture}
  \begin{axis}[
    scale only axis,
    xmin=-0.5,xmax=4.4,
    ymin=82, ymax=87,
    axis y line*=left,% the ’*’ avoids arrow heads
    xlabel=Ratio of vanilla and adversarial training,
    ylabel=Robust Accuracy]
    % \addplot [blue, solid, mark=square, mark options={solid}, mark size=3pt]     
    \addplot + [sharp plot] coordinates {
      (0,86.10)(1,85.52)(2,85.46)(3,83.93)(4,83.27)
      };
    \label{plot_one}
  \end{axis}
  \begin{axis}[
    scale only axis,
    xmin=-0.5,xmax=4.4,
    ymin=16,ymax=55,
    axis y line*=right,
    axis x line=none,
    ylabel=Training time(in hours)]
    \addplot[red, mark=triangle*] [sharp plot] coordinates {(0,49.7)(1,33.5)(2,25.1)(3,21.96)(4,19.5)};
    \label{plot_two}
    \addlegendimage{/pgfplots/refstyle=plot_one}\addlegendentry{Training Time}
    \addlegendimage{/pgfplots/refstyle=plot_two}\addlegendentry{Robust Accuracy}
    %\legend{Training Time}
  \end{axis}
\end{tikzpicture}
}

\caption{Graph showing the tradeoff between robust accuracy and training time by changing ratio between vanilla and adversarial training on the CIFAR-10 dataset. As the proportion of vanilla training increases the training time decrease significantly while decreasing the robust accuracy marginally.}
\label{frequency}
\end{figure}

\subsection{Results on MNIST}
We tested our approach on the MNIST dataset using FGSM adversarial training. In Table \ref{tab:results_tab} we show the results of only adversarial training and our approach. For our experiments, we used standard hyperparameters used by Madry \cite{b13}. During training the adversarial examples are constructed with \(\epsilon\) = 0.3 as suggested in \cite{b13}. During the test time, \(\epsilon\) = 0.3 is used to measure robust accuracy. Using our novel approach of combining adversarial training with vanilla training, we achieved a 3.52 speed up in training time, while increasing the robust accuracy marginally from 97.71\% to 98.71\%. We deduce that the robust accuracy increased on MNIST due to better adversarial example generation from the adversarially-prone subset.

% \vipul{Para2} 
\subsection{Results on CIFAR-10}
Next, we evaluated our approach on the CIFAR-10 dataset using FGSM adversarial training. We trained the CIFAR-10 model using Wide-Resnet 32-10 model and standard hyperparameters used by Madry \textit{et al.} \cite{b13}. During training the adversarial examples are constructed with \(\epsilon\) = 0.0157 (4/255) as suggested by \cite{b13}. Same \(\epsilon\) value is used to measure robust accuracy during test time. The results are shown in Table \ref{tab:results_tab}. Conventional adversarial training on the CIFAR-10 dataset took around 50 hours to converge. By using our approach, we were able to decrease the training time by a factor of 1.98 and bring it down to 25 hours. The robust accuracy of the model decreased marginally by 0.64\%. This shows that we were able to achieve a similar level of robustness by using a subset of training data. We propose that for tasks where training time ranges in days or weeks, our approach can be very effective. A speed of around 2 will result in saving days and weeks of training time and computation.

\subsection{Results on Fast-adv}
Next, we evaluated our approach on a state-of-the-art Free adversarial training algorithm \cite{b24} on the ImageNet dataset. We used the open-sourced code for Free adversarial training and performed experiments for 50 epochs on the ImageNet dataset. The value of \(\epsilon\) = 2/255 for adversarial example generation and testing. As shown in Table \ref{tab:results_tab}, we achieved a speedup of 1.2x, thus decreasing the training time by more than 10 hours. The robust accuracy decreases marginally by 1.68\% from 83.52 to 81.84. We report top5 precision accuracy on the test set. We deduce that training time decreased significantly as we save time on generating adversarial attacks on the entire training dataset and use only a subset of it for adversarial training. This shows that our approach can be applied to state-of-the-art adversarial training techniques and it will help in decreasing training time.

% \vipul{Para4} For all results we perform the training multiple times and report the average training time numbers. GPU used info

We performed the experiments four times to normalize the effect of randomness. The reported accuracy and training time are the average numbers for all experiments. We have performed all the experiments on CentOS. We used 2 and 4 NVIDIA Tesla V100 16GB GPUs for FGSM adversarial training on MNIST and CIFAR-10 datasets respectively. For Free adversarial training, we used 4 NVIDIA Tesla V100 16GB GPUs.

% \vipul{Para5} Tradeoff between accuracy and time

In Figure \ref{frequency}, we show the change in robust accuracy and training time with the change in the ratio of vanilla and adversarial training on CIFAR-10 dataset. A ratio of 0 is equivalent to performing only adversarial training, without mixing it with vanilla training. The training time decreases significantly as we increase the proportion of vanilla training. This is accompanied by a decrease in robust accuracy. We found that using a ratio of 2 gives the best tradeoff between robust accuracy and training time. This means that we perform adversarial training on every 3rd iteration on the adversarially-prone subset of training data.

\section{Conclusion}

 Recent approaches in adversarial training have made great progress in improving the robustness of deep neural networks. Though improving the robustness of the model, adversarial training algorithms increase the computation time drastically. We show that we don't have to use the entire training dataset for adversarial training. Adversarial training, performed on most adversarially prone samples from the training dataset, can be combined with vanilla training to decrease the training time. Our approach can be plugged into any adversarial training algorithm. We show that after plugging in our approach, adversarial training algorithms have a speed of at least 1.2x, with a marginal drop in robust accuracy.

\bibliography{references}

\begin{thebibliography}{00}
\bibitem{b1} Ken Kingery, "Stop Gambling with Black Box and Explainable Models on High-Stakes Decisions" , 2019.

\bibitem{b2} Alex Krizhevsky, Ilya Sutskever and Geoffrey E Hinton. Imagenet classification with deep convolutional neural networks. In Advances in neural information processing systems, 2012, pp. 1097-1105.

\bibitem{b3} Ronan Collobert and Jason Weston. A unified architecture for natural language processing: Deep neural networks with multitask learning.
In Proceedings of the 25th international conference on Machine learning(ACM), 2018, pp. 160-167.

\bibitem{b4} Håvard Kvamme, "Credit Scoring with Deep Learning", 2020

\bibitem{b5} Ihor Starepravo, "Super-Weak AI: Evolution Dead-End or Future of Autonomous Driving?", 2018

\bibitem{b6} Mingyu Kim, "Deep Learning in Medical Imaging", 2020

\bibitem{b7} Miles Brundage, Shahar Avin, Jasmine Wang, Yoshua Bengio, et al. Toward Trustworthy AI Development: Mechanisms for Supporting Verifiable Claims. \textit{arXiv preprint arXiv:2004.07213, 2020}

\bibitem{b8} Cynthia Rudin. Stop Explaining Black Box Machine Learning Models for High Stakes Decisions and use Interpretable Models Instead. Nature Machine Intelligence, 2019 

\bibitem{b9} Ian J. Goodfellow, Jonathon Shlens and Christian Szegedy. Explaining and Harnessing Adversarial Examples.
    In \textit{International Conference on Learning Representations(ICLR), 2015}

\bibitem{b10} Seyed-Mohsen, Moosavi-Dezfooli, Alhussein Fawzi and Pascal Frossard. DeepFool: A Simple and Accurate Method to Fool Deep Neural Networks. In \textit{Computer Vision and Pattern Recognition(CVPR), 2016}

\bibitem{b11} Christian Szegedy, Wojciech Zaremba, Ilya Sutskever, Joan Bruna, Dumitru Erhan, Ian Goodfellow and Rob Fergus. Intriguing properties of neural networks. In \textit{International Conference on Learning Representations, 2014}

\bibitem{b12} Battista Biggio, Igino Corona, Davide Maiorca, Blaine Nelson, Nedim Srndic, Pavel Laskov, Giorgio Giacinto and Fabio Roli. "Evasion Attacks against Machine Learning at Test Time". In ECML/PKDD 2013, pp. 387-402 

\bibitem{b13} Aleksander Madry, Aleksandar Makelov, Ludwig Schmidt,  Dimitris Tsipras and Adrian Vladu. Towards deep learning models resistant to adversarial attacks. In \textit{International Conference on Learning Representations, 2018}

\bibitem{b14} Xiaoyong Yuan, Pan He, Qile Zhu and Xiaolin Li. Adversarial Examples: Attacks and Defenses for Deep Learning. In \textit{IEEE Transactions on Neural Networks and Learning Systems, 2019}

\bibitem{b15} Anish Athalye, Nicholas Carlini and David Wagner. Obfuscated Gradients Give a False Sense of Security: Circumventing Defenses to Adversarial Examples. In Proceedings of the 35th International Conference on Machine Learning, 2018.

\bibitem{b16} Alexey Kurakin, Ian Goodfellow, Samy Bengio, Yinpeng Dong, Fangzhou Liao, Ming Liang, Tianyu Pang and et al. Adversarial Attacks and Defences Competition. In The NIPS '17 Competition: Building Intelligent Systems

\bibitem{b17} Andras Rozsa, Ethan M. Rudd and Terrance E. Boult. Adversarial Diversity and Hard Positive Generation. In \textit{Proceedings of the IEEE Conference on Computer Vision and Pattern Recognition (CVPR) Workshops, 2016}, pp. 25-32

\bibitem{b18} S. M. Moosavi-Dezfooli, A. Fawzi, O. Fawzi and P. Frossard. Universal adversarial perturbations. In \textit{Proceedings of IEEE Conference on Computer Vision and Pattern Recognition (CVPR), 2017}

\bibitem{b19} Tao Xiang, Hangcheng Liu, Shangwei Guo, Tianwei Zhang and Xiaofeng Liao. Local Black-box Adversarial Attacks: A Query Efficient Approach.      \textit{arXiv preprint arXiv:2101.01032, 2021}

\bibitem{b20} Nicolas Papernot, Patrick McDaniel, Somesh Jha, Matt Fredrikson, Z. Berkay Celik and Ananthram Swami. "The Limitations of Deep Learning in Adversarial Setting". In 2016 IEEE European Symposium on Security and Privacy (EuroS P), 2016. pp 372-387

\bibitem{b21} Alexey Kurakin, Ian Goodfellow and Samy Bengio. Adversarial Machine Learning at Scale. In \textit{International Conference on Learning Representations (ICLR), 2017}

\bibitem{b22} Maosen Li, Cheng Deng, Tengjiao Li, Junchi Yan, Xinbo Gao and Heng Huang. Towards Transferable Targeted Attack. In \textit{ Proceedings of the IEEE/CVF Conference on Computer Vision and Pattern Recognition (CVPR), 2020}

\bibitem{b23} Wenjie Wan, Zhaodi Zhang, Yiwei Zhu, Min Zhang and Fu Song. Accelerating Robustness Verification of Deep Neural Networks Guided by Target Labels. \textit{arXiv preprint arXiv:2007.08520, 2020}

\bibitem{b24} Ali Shafahi, Mahyar Najibi, Mohammad Amin Ghiasi, Zheng Xu, John P Dickerson, Christoph Studer, Larry Davis, Gavin Taylor and Tom Goldstein. Adversarial training for free!. In \textit{Neural Information Processing Systems(NeurIPS), 2019}

\bibitem{b25} Eric Wong, Leslie Rice, J. Zico Kolter. Fast is better than free: Revisiting adversarial training. In \textit{International Conference on Learning Representations (ICLR), 2020}

\bibitem{b26} Shumeet Baluja and Ian Fischer. Adversarial transformation networks: Learning to generate adversarial examples. In \textit{Association for the Advancement of Artificial Intelligence (AAAI), 2018}

\bibitem{b27} Yanpei Liu, Xinyun Chen, Chang Liu and Dawn Xiaodong Song. Delving into Transferable Adversarial Examples and Black-box Attacks. In \textit{arXiv preprint arXiv:1611.02770, 2016}

\bibitem{b28} Reuben Feinman, Ryan R. Curtin, Saurabh Shintre and Andrew B. Gardner. Detecting Adversarial Samples from Artifacts. In \textit{arXiv preprint arXiv:1703.00410, 2017}

\bibitem{b29} Shiqi Wang, Kexin Pei, Justin Whitehouse, Junfeng Yang and Suman Jana. Formal Security Analysis of Neural Networks using Symbolic Intervals. \textit{arXiv preprint arXiv:1804.10829, 2018}

\bibitem{b30} Shiqi Wang, Kexin Pe, Justin Whitehouse, Junfeng Yang and Suman Jana. Efficient Formal Safety Analysis of Neural Networks. In \textit{Neural Information Processing Systems(NeurIPS), 2018}

\bibitem{b31} Weilin Xu, David Evans and Yanjun Qi. Feature Squeezing: Detecting Adversarial Examples in Deep Neural Networks \textit{arXiv preprint arXiv:1704.01155, 2018}

\bibitem{b32} Yigit Alparslan, Ken Alparslan, Jeremy Keim-Shenk, Shweta Khade and Rachel Greenstadt. Adversarial Attacks on Convolutional Neural Networks in Facial Recognition Domain. \textit{arXiv preprint arXiv:2001.11137, 2020}

\bibitem{b33} Jiawei Su, Danilo Vasconcellos Vargas and Kouichi Sakurai. One Pixel Attack for Fooling Deep Neural Networksvol . In \textit{IEEE Transactions on Evolutionary Computation, 2019}

\bibitem{b34} Chenchen Zhao and Hao Li. Blurring Fools the Network -- Adversarial Attacks by Feature Peak Suppression and Gaussian Blurring. \textit{arXiv preprint arXiv:2012.11442, 2020}

\bibitem{b35} Rafael Pinot, Raphael Ettedgui, Geovani Rizk, Yann Chevaleyre and Jamal Atif. Randomization matters. How to defend against strong adversarial attacks. In \textit{International Conference on Machine Learning(ICML), 2020}
   
\bibitem{b36} Gaurav Goswami, Nalini Ratha, Akshay Agarwal, Richa Singh and Mayank Vatsa. Unravelling Robustness of Deep Learning based Face Recognition Against Adversarial Attacks. In \textit{Proceedings of the AAAI Conference on Artificial Intelligence, 2018}

\bibitem{b37} Jan Hendrik Metzen, Tim Genewein, Volker Fischer and Bastian Bischoff. On Detecting Adversarial Perturbations. \textit{arXiv preprint arXiv:1702.04267, 2017}

\bibitem{b38} Chaowei Xiao, Jun-Yan Zhu, Bo Li, Warren He, M. Liu and Dawn Xiaodong Song. Spatially Transformed Adversarial Examples. \textit{arXiv preprint arXiv:1801.02612, 2018}

\bibitem{b39} Nicholas Carlini and David A. Wagner. Towards Evaluating the Robustness of Neural Networks. \textit{2017 IEEE Symposium on Security and Privacy (SP), 2017}

\bibitem{b40} Nicholas Carlini and David A. Wagner. Adversarial Examples Are Not Easily Detected: Bypassing Ten Detection Methods. \textit{Proceedings of the 10th ACM Workshop on Artificial Intelligence and Security, 2017}

\bibitem{b41} Nicholas Carlini, Anish Athalye, Nicolas Papernot, Wieland Brendel, Jonas Rauber, Dimitris Tsipras, Ian J. Goodfellow, Aleksander Madry and Alexey Kurakin. On Evaluating Adversarial Robustness. \textit{arXiv preprint arXiv:1902.06705, 2019}

\bibitem{b42} Yinpeng Dong, Qi-An Fu, Xiao Yang, Tianyu Pang, Hang Su, Zihao Xiao and Jun Zhu. Benchmarking Adversarial Robustness. \textit{arXiv preprint arXiv:1912.11852, 2019}

\bibitem{b43} Yinpeng Dong, Qi-An Fu, Xiao Yang, Tianyu Pang, Hang Su, Zihao Xiao and Jun Zhu. Benchmarking Adversarial Robustness. \textit{IEEE/CVF Conference on Computer Vision and Pattern Recognition (CVPR), 2020}

\bibitem{b44} Fangzhou Liao, Ming Liang, Yinpeng Dong, Tianyu Pang, Jun Zhu and Xiaolin Hu. Defense Against Adversarial Attacks Using High-Level Representation Guided Denoiser. \textit{IEEE/CVF Conference on Computer Vision and Pattern Recognition (CVPR), 2018}

\bibitem{b45} Mathias L{\'e}cuyer, Vaggelis Atlidakis, Roxana Geambasu, Daniel J. Hsu and Suman Sekhar Jana. Certified Robustness to Adversarial Examples with Differential Privacy. \textit{IEEE Symposium on Security and Privacy (SP), 2019}

\bibitem{b46} Jeremy M. Cohe, Elan Rosenfeld and J. Zico Kolter. Certified Adversarial Robustness via Randomized Smoothing. \textit{International Conference on Machine Learning(ICML), 2019}

\bibitem{b47} Yang Song, Taesup Kim, Sebastian Nowozin, Stefano Ermon and Nate Kushman. PixelDefend: Leveraging Generative Models to Understand and Defend against Adversarial Examples. \textit{arXiv preprint arXiv:1710.10766, 2018}

\bibitem{b48} Chongli Qin, James Martens, Sven Gowal, Dilip Krishnan, Krishnamurthy Dvijotham, Alhussein Fawzi, Soham De, Robert Stanforth and Pushmeet Kohli. Adversarial Robustness through Local Linearization. \textit{Conference on Neural Information Processing Systems, 2019}

\bibitem{b49} Jianyu Wang and Haichao Zhang. Bilateral Adversarial Training: Towards Fast Training of More Robust Models Against Adversarial Attacks. \textit{IEEE/CVF International Conference on Computer Vision (ICCV), 2019}

\bibitem{b50} Vivek B.S. and R. Venkatesh Babu. Single-Step Adversarial Training With Dropout Scheduling. \textit{IEEE/CVF Conference on Computer Vision and Pattern Recognition (CVPR), 2020}

\bibitem{b51} Eric Wong and J. Zico Kolter. Provable defenses against adversarial examples via the convex outer adversarial polytope. \textit{International Conference on Machine Learning(ICML), 2018}

\bibitem{b52} Krishnamurthy Dvijotham, Robert Stanforth, Sven Gowal, Timothy A. Mann and Pushmeet Kohli. A Dual Approach to Scalable Verification of Deep Networks. \textit{arXiv preprint arXiv:1803.06567, 2018}

\bibitem{b53} Vincent Tjeng, Kai Y. Xiao and Russ Tedrake. Evaluating Robustness of Neural Networks with Mixed Integer Programming. \textit{International Conference on Learning Representations (ICLR), 2019}

\bibitem{b54} Huan Zhang, Tsui-Wei Weng, Pin-Yu Chen, Cho-Jui Hsieh and Luca Daniel. Efficient Neural Network Robustness Certification with General Activation Functions. \textit{Conference on Neural Information Processing Systems, 2018}

\bibitem{b55} Vipul Gupta, Zhuowan Li, Adam Kortylewski, Chenyu Zhang, Yingwei Li and Alan Yuille. Swapmix: Diagnosing and regularizing the over-reliance on visual context in visual question
answering. \textit{In Proceedings of the IEEE/CVF Conference on Computer Vision and Pattern
Recognition (CVPR), pages 5078–5088, June 2022. 2, 3}

\bibitem{b56} Vincent Francois-Lavet, Peter Henderson, Riashat Islam, Marc G. Bellemare and Joelle Pineau. An Introduction to Deep Reinforcement Learning. \textit{CoRR, 2018}

\bibitem{b57} Li, Yize, Pu Zhao, Xue Lin, Bhavya Kailkhura, and Ryan Goldh. Less is More: Data Pruning for Faster Adversarial Training arXiv preprint arXiv:2302.12366 (2023).

\bibitem{b58} Bartoldson, Brian R., Bhavya Kailkhura, and Davis Blalock. "Compute-Efficient Deep Learning: Algorithmic Trends and Opportunities." arXiv preprint arXiv:2210.06640 (2022).

\bibitem{b59} Dolatabadi, Hadi M., Sarah Erfani, and Christopher Leckie. l$\infty$-robustness and beyond: Unleashing efficient adversarial training." In Computer Vision–ECCV 2022: 17th European Conference, Tel Aviv, Israel, October 23–27, 2022, Proceedings, Part XI, pp. 467-483. Cham: Springer Nature Switzerland, 2022 

\end{thebibliography}

\vspace{12pt}
% \color{red}
% IEEE conference templates contain guidance text for composing and formatting conference papers. Please ensure that all template text is removed from your conference paper prior to submission to the conference. Failure to remove the template text from your paper may result in your paper not being published.

\end{document}